\documentclass[runningheads]{llncs}
\usepackage{graphicx}
\usepackage{multirow}
\usepackage{booktabs}
\usepackage[caption=false, font=scriptsize]{subfig}
\usepackage[most]{tcolorbox}

\begin{document}

\title{Robustness Analysis of Deep Learning Frameworks on Mobile Platforms}

\author{Amin Eslami Abyane \and
Hadi Hemmati}
\authorrunning{A. Eslami Abyane \and H. Hemmati}
%
\institute{Department of Electrical and Software Engineering, University of Calgary, Canada\\ 
\email{\{amin.eslamiabyane, hadi.hemmati\}@ucalgary.ca}}
\maketitle              
\begin{abstract}
With the recent increase in the computational power of modern mobile devices, machine learning-based heavy tasks such as face detection and speech recognition are now integral parts of such devices. This requires frameworks to execute machine learning models (e.g., Deep Neural Networks) on mobile devices. Although there exist studies on the accuracy and performance of these frameworks, the quality of on-device deep learning frameworks, in terms of their robustness, has not been systematically studied yet.
In this paper, we empirically compare two on-device deep learning frameworks with three adversarial attacks on three different model architectures. We also use both the quantized and unquantized variants for each architecture.
The results show that, in general, neither of the deep learning frameworks is better than the other in terms of robustness, and there is not a significant difference between the PC and mobile frameworks either. However, in cases like Boundary attack, mobile version is more robust than PC. In addition, quantization improves robustness in all cases when moving from PC to mobile.

\end{abstract}
\section{Introduction}
\label{introSec}
In recent years, advancements in hardware resources and demands from many application domains have led to the growth and success of deep learning (DL) approaches. Consequently, several deep learning frameworks, such as  TensorFlow \cite{10.5555/3026877.3026899} and PyTorch \cite{paszke2017automatic}, have been introduced to improve DL developers productivity.
These powerful frameworks have gained massive success and popularity in academia and industry and are being used on a large scale every day.
Among the many application domains of DL systems, one particular domain that has seen much interest is applying DL techniques on mobile devices (i.e., on-device learning). In general, mobile devices' collected or observed data are potentially of great interest for many DL applications such as speech recognition, face detection, and next-word prediction. There are two generic solutions to utilize these data. The first approach is to send the data from the mobile devices to a server to run the DL task (training or testing) and return the results.
This approach has some significant drawbacks. The first one is that in this communication with the server, the user's privacy might be threatened, and the second one is that we are adding a network overhead and delay to the system.
This overhead might get very noticeable if the task is frequent, such as image classification using a camera to process lots of images in a second.

The second approach is called on-device machine learning, which does not have the privacy concerns and is the context of this paper.
In an on-device machine learning approach, the inference is made on the user's device, and no network communication is required. However, the mobile device's computational power (no matter how powerful the mobile device is) is much less than a server or even a regular GPU-based PC, which is an obstacle for training in this fashion.
To help DL models inference possible on a mobile device, libraries such as TensorFlow Lite\cite{TensorFl80:online} and PyTorch Mobile\cite{HomePyTo69:online} have been proposed.

From the software engineering perspective, an important quality aspect of DL-based software systems is their robustness, which is usually tested and analyzed against adversarial attacks \cite{7958570}, \cite{8601309}, \cite{10.5555/3157096.3157279}. Robustness is especially significant in mobile apps given the amount of personal information that can be exploited from cell phones, if an adversary gets access to the app. For instance, DL-based face detection is a standard access control measure for cell phones these days. Suppose an adversarial attack on the underlying DL model can misclassify a specific image created by the adversary as the trusted class. In that case, the attacker gets access to the mobile device.

There have been some limited studies in the recent literature that evaluate DL frameworks both on PC and mobile \cite{DBLP:conf/eccv/IgnatovTCWWHG18}, \cite{DBLP:conf/kbse/GuoCXMHLLZL19}, \cite{DBLP:journals/corr/abs-2005-05085}, but only in terms of their accuracy (effectiveness) and performance (efficiency). This has motivated us to conduct this study with a software testing and analysis lens, on mobile and PC DL frameworks in terms of their robustness.

We study two main on-device DL frameworks, TensorFlow Lite and PyTorch Mobile, from Google and Facebook, respectively. We use image classification as a common DL task to evaluate the robustness of the frameworks. Our controlled experiment is designed to study the effect of the models, the adversarial attacks, the quantization process \cite{jacob2017quantization}, and the framework on robustness. We compare two deep learning frameworks (TensorFlow and PyTorch) with three adversarial attacks (both white-box and black-box) on three different model architectures, both quantized and unquantized. This results in 36 configurations on mobile devices and 18 configurations on PC, as our comparison baseline.

The results show that neither of the mobile deep learning frameworks is better than the other in terms of robustness, and the robustness depends on the model type and other factors, which is the case on PC as well. Moreover, there is no significant difference in robustness between PC and mobile frameworks either. However, cases like the Boundary attack on PyTorch, we see that the mobile version is significantly more robust than the PC version (12.5\% decrease in attack success rate). Finally, we see that quantization improves the robustness of both TensorFlow and PyTorch on all models and attacks when moving from PC to mobile (with median improvements between 2.4\% to 23.8\%, per attack type). Note that all data and scripts are available for replication purposes \footnote{https://github.com/aminesi/robustness-on-device}.


\section{Background}
\label{backSec}

\subsection{Deep Neural Network (DNN)}
\label{DNN}
A DNN is an artificial neural network consisting of many layers, and each layer has multiple neurons. Each neuron performs a simple task, takes an input, and gives an output based on a function.
A simple combination of these layers is often called multi-layer perceptron (MLP). However, DNNs are not limited to MLPs. 
One of the most popular kinds of neural networks is called Convolutional Neural Networks (CNNs). A convolutional layer is typically used in tasks that work with images.
A convolutional layer's objective is to extract features from a picture and reduce the problem's dimensionality.
Another group of neural networks is called Recurrent Neural Networks (RNNs).
This type of network has units that act as memory. Thus they are often used in tasks that deal with language and speech.
Like a human being, DNN learns patterns in the training phase and can be used for the designed task.
DNNs are extremely powerful and are widely used for image classification, face recognition, and speech recognition.


In the image classification domain, which is a common application domain of DNNs and is the domain of our experiment, some very well-known models have proven to be very effective:

\textbf{MobileNetV2}\cite{DBLP:conf/cvpr/SandlerHZZC18}: 
This model is specially designed for mobile devices and is more light-weight than the other models. It contains 32 convolutional filters and 19 layers of residual bottleneck. The overall size of this model is around 14 MB.
This model takes images of size 224×224 as input for classification.

\textbf{ResNet50}\cite{DBLP:conf/cvpr/HeZRS16}: 
This is one of the most influential models in the image classification domain. 
It is much heavier than MobileNetV2 (it is close to 100 MB), but it is more accurate than MobileNetV2.

\textbf{InceptionV3}\cite{DBLP:conf/cvpr/SzegedyVISW16}: 
Much like ResNet50, this model is another complex model (close to 100 MB).
It consists of inception blocks; 
Unlike the other two models, which used 224×224 images, this one takes images of size 299×299, which is one reason it is more complex than the previous ones.

\subsection{Robustness and Adversarial Attacks}
\label{Attacks}
DNNs are known to be sensitive to corner case data (e.g., data items that are close to decision boundaries). That means it is possible that a slight change in the input can result in a corner case sample where the DNN will not perform accurately (e.g., the item is misclassified if the task is classification). 
Suppose this slight change of input is deliberate to fool the model. In that case, it is called an adversarial attack, and the robustness of a DNN model is the extent that the model can defend itself from such attacks (i.e., still generate correct outputs).
Adversarial attacks were first introduced in the image processing tasks, where images are easily manipulable, and the tasks (e.g., classification) are pretty sensitive. However, these attacks have gone beyond the image domain and are now being studied in other learning tasks, such as text and audio analysis domains.

Adversarial attacks can be categorized from several perspectives.
Most commonly, they are categorized into two groups, based on their level of access to the model details, which are white-box and black-box attacks.
White-box attacks require knowledge about the internals of the models they are attacking. For instance, some attacks need the models' gradient after the backpropagation step to generate adversarial samples, whereas black-box attacks do not need such information and are model-agnostic.
Another standard categorization of attacks is grouping them into targeted and untargeted attacks.
Targeted attacks try to fool the model into misclassifying data into a particular class. In contrast, untargeted attacks just try to force the model to misclassify, no matter the wrong output. Since most of the popular and well-known attacks are defined as untargeted \cite{DBLP:journals/corr/GoodfellowSS14}, \cite{kurakin2017adversarial}, \cite{7780651}, in this study we focus on the following untargeted attacks both from white-box and black-box categories.



\textbf{Fast Gradient Sign Method (FGSM)}
     \cite{DBLP:journals/corr/GoodfellowSS14} is perhaps the most famous adversarial attack amongst these attacks. 
    FGSM is a gradient-based white-box attack, and it works by adding perturbations into the image following the formula presented in equation \ref{fgsm-formula}.
    Where x and y are the input image and label respectively, $\Theta$ represents model parameters, $\nabla$ is gradient, J is the loss function, and $\epsilon$ is the amount of perturbation.
          \begin{equation}
              adv = x+\epsilon.sign(\nabla(J(\theta, x, y)))
              \label{fgsm-formula}
          \end{equation}

\textbf{Basic Iterative Method (BIM)}
     \cite{kurakin2017adversarial} is an extension of FGSM attack, so it is also a white-box attack.
          As its name suggests, it is iterative, and it does the FGSM attack multiple times and updates the input in each iteration.

\textbf{Boundary Attack} \cite{DBLP:conf/iclr/BrendelRB18}  is a decision-based (it only uses the final decision of model to create samples) black-box attack.
    It starts from an adversarial point, and in each step makes a move randomly orthogonal to the image and a move towards the image.
    Since it uses a small step size to get closest to the image (while staying on the adversarial side), it requires many iterations.

\subsection{DL Frameworks}
Neural networks use complex mathematical equations that need to be implemented in
Libraries like TensorFlow and PyTorch that provide a set of simple Application Programming Interfaces (APIs) for the machine learning developers. Both these frameworks are implemented in python language for their high-level APIs, and they use C++ for their low-level implementation to gain higher speeds.

With the advances in on-device learning, these frameworks are now coming with mobile versions, making programming DL on mobile devices easier.

TensorFlow's mobile variant called ``TensorFlow Lite'' is a cross-platform (Android, iOS, and edge devices) library and supports languages such as Java and Swift, which is based on a cross-platform serialization library called FlatBuffers. In addition, it supports multiple quantization configurations and different hardware such as CPU and GPU with various options, and on Android, it supports Android Neural Networks API (NNAPI).

``PyTorch Mobile'' is the other on-device DL library, which is similar to TensorFlow Lite in terms of functionalities. It works on the same platforms that TensorFlow Lite does and supports quantization but is less flexible in this aspect. At the time of doing this experiment, PyTorch Mobile only supports CPU without any additional options (e.g., thread count).

However, like any other software program, these implementations are not flawless. There are also quite many different design choices and implementation differences between various frameworks. Especially, given the hardware and operating systems differences, there might be many variations in terms of effectiveness, efficiency, and robustness of the same model implemented on PC vs. mobile in different frameworks.

\subsection{Quantization}
\label{quantization}
Quantization is the process of compressing a DNN implementation to speed up the model execution, at the cost of its precision \cite{Modelopt44:online}. As we know, DNN models' implementations in DL frameworks include many matrices/tensors and thus many matrix/tensor operations.
The motivation behind quantization is to reduce the complexity of matrix operations to be able to run more operations with fewer resources \cite{DBLP:journals/corr/abs-2004-09602}.
Typically, all DNN model parameters, like weights and activations, use a 32-bit floating-point precision.
Since mobile devices have fewer resources than PCs, the quantization idea has been proposed to slightly reduce the model's precision to make models smaller and infer faster.
The quantization target may be an 8-bit integer, or 16-bit floating-point, or any other precision for the numerical data types.

Integer quantization is explored in \cite{jacob2017quantization}. In this work, they introduce a formula for quantization to 8-bit integer, which is shown in equation \ref{quant-formula}. Key parameters here are zero\_point and scale, which should be selected in a way that every possible normal\_value can be mapped to an 8-bit fixed-point value. 

\begin{equation}
    normal\_value=(int8\_value-zero\_point) \times scale
    \label{quant-formula}
\end{equation}
          
Although TensorFlow currently supports multiple precisions, PyTorch only supports an 8-bit integer at the moment.
There are different ways of quantization, and we will briefly describe them in this section.

\textbf{Post-training dynamic range quantization:}
This type of quantization \cite{Introduc74:online}, as the name suggests, is done after the model is trained.
A quantizer quantizes all statically defined parameters, like weights using a formula like equation \ref{quant-formula}.
So in this approach, the activations are intact. According to \cite{Introduc74:online}, this approach works best for language models.
The name dynamic is used since the activation values are quantized on the fly when the model is running.

\textbf{Post-training static range quantization:}
This quantization approach \cite{Introduc74:online}, like the previous one, is applied on a trained model.
The difference is that activations are pre quantized as well, making the model more compact and faster.
To find the best quantization, the quantizer should find the best scale and zero\_point to ensure successful mapping to the target (e.g., int8), which is not possible for something that is not statically defined like activations.

The solution is to calibrate the model with some image samples in the quantization step so that quantizer can see various possible dynamic values (activation values in this case) and calculate appropriate scale and zero\_points for them.
This approach is more appropriate for image models compared to the previous one as suggested by the literature \cite{Introduc74:online}.

\textbf{Quantization aware training:}
This approach \cite{Introduc74:online} is distinct from the other two in that it tries to learn the effect of quantization in the training phase. It is a more robust approach but costs more.

\section{Experiments}
\label{experimentSec}

Our objective is to quantitatively and systematically evaluate the robustness of DL learning frameworks on mobile. 
To address this objective, we answer the following research questions:

\begin{itemize}
    \item \textbf{RQ1: How robust DL frameworks are on mobile?}
    This RQ aims to compare TensorFlow Lite and PyTorch Mobile when running on mobile by assessing their robustness against well-known adversarial attacks.
  
    \item \textbf{RQ2: How does mobile DL frameworks' robustness compare to their PC equivalent?}
    In this RQ, we compare the robustness results of DL frameworks on PC vs. mobile platforms.
 
    \item \textbf{RQ3: What is the effect of quantization on the robustness of models?}
    In this RQ, we will study the quantization's effect by repeating the experiment designed for RQ1, but this time with the quantized models. 
\end{itemize}

\subsection{Experiment Design}

\textbf{Models under study and the datasets:}
Image classification is one of the main application domains of DL models these days. Especially in the mobile application domain, some use cases such as access control, as discussed earlier, are critical and can be heavily dependant on the trustworthiness of the underlying classification models.
Therefore, in this study, we focus on image classification and use the three well-known image classification models explained in Section~\ref{DNN}: MobileNetV2, ResNet50, and InceptionV3. These models have been selected to cover various models in terms of power and complexity (MobileNetV2: light-weight, ResNet50: resource-demanding, and InceptionV3: the heaviest).

For TensorFlow models, we use pre-trained models on ImageNet from Keras applications \cite{KerasApp64:online}. 
PyTorch models are all from torchvision \cite{torchvis33:online} models packages
(for the sake of quantization, models are selected from quantization packages as regular models are not quantizable).

For quantization, we use the second approach explained in Section~\ref{quantization}, post-training static range quantization, which is a decent fit for image data and still not very costly.
For calibration of the models in the quantization mode, we use 1,000 random samples from our dataset.

We use ImageNet \cite{ImageNet5:online} as our main dataset.
ImageNet is one of the biggest and most well-known datasets in the image domain, and it uses WordNet \cite{WordNetA96:online} for its label hierarchy.
It consists of 1,281,167 training images and 50,000 validation images with a total of 1,000 image classes. It is around 150 gigabytes in size.

Since our robustness analysis heavily depends on the ground truth for classification, we need to make sure that the original test set samples are correctly classified.
By doing this, we ensure that if an adversarial sample is misclassified, it is due to perturbations and not intrinsic model errors.

To achieve this goal, we use the intersection of correctly classified validation samples by all the models on all of our frameworks.
Then we choose 3,000 samples from these, randomly, to ensure we have enough unbiased samples.
        \vspace{0.01\linewidth}

\textbf{Model deployment and inference:}
The procedure to deploy and test our models on mobile frameworks is as follows:
\begin{enumerate}
    \item Create a trained model on PC (by creating a new model or using a pre-trained model or fine-tune a pre-trained models with transfer learning)
    \item Convert the models into their mobile variant, using TensorFlow Lite and PyTorch Mobile (optionally quantize the model (only in RQ2)).
    \item Load the model and samples into memory and run the inference on mobile to calculate the robustness.
\end{enumerate}

Recall that this procedure is divided between PC (model training) and mobile (model inference), since mobile devices alone are not powerful enough to do heavy tasks such as training an extremely resource-intensive neural network.
        \vspace{0.01\linewidth}

\textbf{Adversarial Attacks:}
As discussed in Section \ref{Attacks}, we use the following three famous untargeted attacks in this study: FGSM, BIM, and Boundary attack. We utilize a python package for creating adversarial attacks called Foolbox \cite{rauber2017foolboxnative}.
For FGSM and BIM attacks, we use an $\epsilon$ of 0.005. For BIM attack, we choose ten iterations which is the default value in Foolbox. 
For Boundary attack, both orthogonal and towards steps are set to 0.01 (which are again the default values in Foolbox), and the number of steps is set to 5000. A small number of steps results in a very perturbed image, and a huge one is highly time-consuming and does not always result in a better sample.

\begin{figure}[t]
    \vspace{-0.01\linewidth}
    \centering
    \subfloat[Original]{\includegraphics[width=0.2\linewidth]{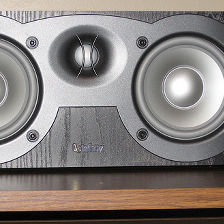}}
    \hspace{0.05\linewidth}
    \subfloat[FGSM]{\includegraphics[width=0.2\linewidth]{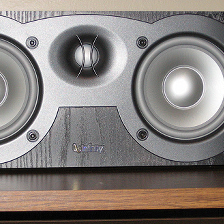}}
    \hspace{0.05\linewidth}
    \subfloat[BIM]{\includegraphics[width=0.2\linewidth]{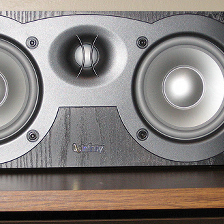}}
    \hspace{0.05\linewidth}
    \subfloat[Boundary]{\includegraphics[width=0.2\linewidth]{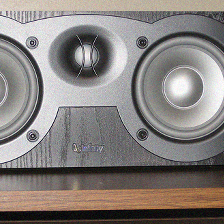}}
    \caption{A sample generated for TensorFlow MobileNetV2 model.}
    \label{attacks}
    \vspace{-0.045\linewidth}

\end{figure}

To generate adversarial samples, we follow these steps that produces visually acceptable images (see \figurename{\ref{attacks}}):

\begin{enumerate}
    \item Preprocess our carefully selected samples according to the model requirements, as follows:
        First Resize the image's smallest dimension to 256 (299 in case of InceptionV3) (This is the choice of both TensorFlow, and PyTorch  \cite{torchvis36:online}, \cite{Inceptio34:online}).
        Then center crop the resized image according to input size of the model.
        Afterwards for PyTorch, divide image pixel values by 255 to map the values to [0, 1] range (TensorFlow uses (0, 255) range).
        For ResNet50 only in TensorFlow, change image format from RGB to BGR. 
        Finally normalize the image channels based on the mean and standard deviation, in Table \ref{normalizaion}, which is from TensorFlow and PyTorch documentations \cite{kerasapp55:online}, \cite{torchvis36:online}.
    \item Pass the model and samples to an attacker to generate adversarial samples.  
    \item Convert adversarial samples to PNG images to use in mobile devices.
\end{enumerate}

\begin{table}[t]
    \setlength\tabcolsep{5pt}
    \caption{Input image normalization in preprocessing. (values are according to image channels (B, G, R) for ResNet50 in TF and (R, G, B) for other configurations)}
    \label{normalizaion}
    \centering
    \begin{tabular}{cccc}
    \bottomrule
    Framework & model &  mean ($\mu$)  & std ($\sigma$) \\
    \toprule
    \multirow{3}{*}{TensorFlow} & MobileNetV2 & (127.5, 127.5, 127.5)& (127.5, 127.5, 127.5)\\
    & ResNet50 & (103.939, 116.779, 123.68) & (1, 1, 1)\\
    & InceptionV3 & (127.5, 127.5, 127.5) & (127.5, 127.5, 127.5)\\
    \hline
    \multirow{3}{*}{PyTorch} & MobileNetV2 & (0.485, 0.456, 0.406) & (0.229, 0.224, 0.225)\\
    & ResNet50 & (0.485, 0.456, 0.406) & (0.229, 0.224, 0.225)\\
    & InceptionV3 & (0.485, 0.456, 0.406) & (0.229, 0.224, 0.225)\\
    \bottomrule
    \end{tabular}

\end{table}

Note that we do not generate the adversarial samples on mobile for two reasons: (a) the computation intensity of this task, and (b) models on mobile do not provide essential information for white-box attacks.
Another point is that we convert images to PNG format, which is a lossless format.
This format is crucial since we want our image to be the same as one on PC to be able to get reliable results.
Also, note that for the Boundary attack, we may need to rerun the algorithm several times until all adversarial samples are successfully generated (in our case, we had to rerun three times).
        \vspace{0.01\linewidth}

\textbf{Evaluation metrics:}
To evaluate robustness, we use success rate, which measures the proportion of samples that could successfully fool the models.
Since our originally selected test set samples are all classified correctly on all configurations on PC, the success rate of an attack has an inverse correlation with the model's robustness.

In RQ2, to better assess the differences between results, when comparing success rates of quantized vs unquantized models, we run a non-parametric statistical significant test (Wilcoxon signed-rank) and report the effect size measure (Vargha and Delaney A Measure), as well.
        \vspace{0.01\linewidth}

\textbf{Execution environment}
We run our PC DL frameworks on a node from the Compute Canada cluster with 32 gigabytes of RAM, an Nvidia V100 GPU, and an Intel Gold 6148 Skylake @ 2.4 GHz CPU. 
We use a physical device for our mobile device, an HTC U11 with a Qualcomm Snapdragon 835 chipset and 6 gigabytes of RAM, running Android 9. We did not use multiple devices since model robustness is independent of device type, given that implementations are using the same library. 
It is worth noting that we do not use an emulator as our mobile system. 
We started the project by testing TensorFlow with an emulator, but surprisingly, we found that quantized models ran slower than regular models in the emulator.
It turned out that TensorFlow's quantized kernel is only optimized for mobile CPU.
Thus quantized versions ran poorly on an emulator.
Therefore, we used a real device, as discussed before.
We developed a prototype mobile app (on both frameworks) that takes configuration and images as input and calculates the success rate .

\subsection{Results and Discussions}
In this section, we present and discuss the results of RQ1 to RQ3.

\begin{figure}[t]
    \centering
    \hspace{-0.045\linewidth}
    \subfloat[MobileNetV2\label{android_robustness_a}]{%
        \includegraphics[width=0.375\linewidth]{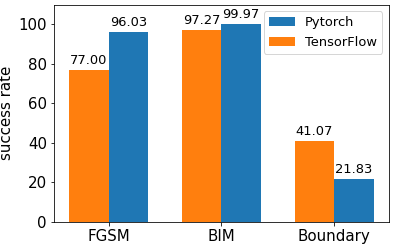}}
    \subfloat[ResNet50\label{android_robustness_b}]{%
        \includegraphics[width=0.33\linewidth]{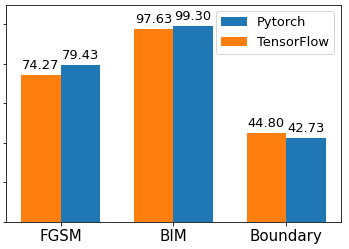}}
    \subfloat[InceptionV3\label{android_robustness_c}]{%
        \includegraphics[width=0.33\linewidth]{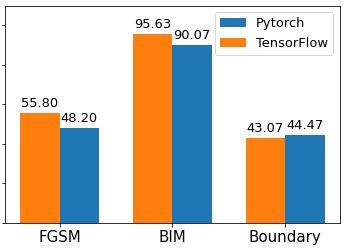}}
    \caption{Success rate of adversarial attacks on mobile device}
    \label{android_robustness}
\end{figure}

\textbf{RQ1 results (TensorFlow Lite vs. PyTorch Mobile robustness on Mobile platforms):}
\figurename{\ref{android_robustness}} reports the results for this RQ. 
The first observation is that in MobileNetV2 and ResNet50 models, TensorFlow Lite was more robust against FGSM and BIM attacks (\figurename{\ref{android_robustness_a}} and \figurename{\ref{android_robustness_b}}).
However, in the Boundary attack, PyTorch Mobile was more robust.
Also, \figurename{\ref{android_robustness_c}} shows that, for InceptionV3, PyTorch Mobile is more robust against FGSM and BIM attacks. However, TensorFlow Lite is more robustness against the Boundary attack.

Thus, in mobile DL frameworks, the robustness depends on configurations, and no learning framework among TensorFlow Lite and PyTorch Mobile dominates the other one in terms of robustness.

It can also be seen that the more complex the model gets, the more robust it will be against adversarial samples.
For instance, In TensorFlow Lite, the success rate of the FGSM attack on MobileNetV2, ResNet50, and InceptionV3 is 77\%, 74.27\%, and 55.8\%, respectively.

Furthermore, both FGSM and BIM attacks are much more effective than the Boundary attack. In addition, BIM attack always performs better than FGSM as it tries to improve an FGSM sample iteratively.

Moreover, we can see that Boundary attack is less successful than FGSM and BIM in all cases with a very low success rate. This seems contradictory with the definition of Boundary attack, which was supposed to generate samples always on the adversarial side.
In other words, the success rate should have always been 100\%.
The reason for lower success rates is that images are in 8-bit unsigned integer (UINT8) format, but the neural networks work with 32-bit floating-points (FP32).
Therefore, when a sample is generated, it is in FP32 format and is always adversarial. However, the reduction of precision to UINT8, in conversion to the image format, makes some samples cross the boundary, and the success rate significantly drops.

Besides the attack success rate, another important factor is performance. Table \ref{mobilePerformance} shows the inference time for the mobile platform when running unquantized models. As the table shows in terms of performance (run-time cost), TensorFlow Lite is much faster in all configurations (the slower framework is highlighted in the table) for regular models.
This might be because PyTorch Mobile uses fewer threads as the number of workers cannot be set on PyTorch.

\begin{table}[t]
    \setlength\tabcolsep{5pt}

    \caption{Inference time in the mobile device for the entire test set (3,000 samples). The bold cells represent the framework with faster inferences.}
    \label{mobilePerformance}
    \centering
    \begin{tabular}{cccc}
    \bottomrule
        &   &   \multicolumn{2}{c}{Inference time (s)}\\
        Model  &   Attack &   TensorFlow Lite &   PyTorch Mobile\\
        \toprule
        \multirow{3}{*}{MobileNetV2}    &   FGSM    &   \textbf{157} & 276   \\
                                        &   BIM     &   \textbf{174} & 301   \\
                                        &   Boundary&   \textbf{187} & 302   \\
        \midrule
        \multirow{3}{*}{ResNet50}       &   FGSM    &   \textbf{1003} & 1583   \\
                                        &   BIM     &   \textbf{1036} & 1504   \\
                                        &   Boundary&   \textbf{1093} & 1460   \\
        \midrule
        \multirow{3}{*}{InceptionV3}    &   FGSM    &   \textbf{1617} & 1774   \\
                                        &   BIM     &   \textbf{1626} & 1841   \\
                                        &   Boundary&   \textbf{1516} & 1820   \\
    \bottomrule
    \end{tabular}
            \vspace{-0.02\linewidth}

\end{table}

\textbf{Answer to RQ1:} Neither PyTorch Mobile nor TensorFlow Lite is significantly more robust than the other, in all cases. The choice of a more robustness mobile framework depends on the model architecture and the attack itself. In terms of performance, however, TensorFlow Lite is consistently faster!

\begin{figure}[t]
    \vspace{-0.05\linewidth}
    \centering
    \hspace{-0.045\linewidth}
    \subfloat[MobileNetV2\label{gen_a}]{%
        \includegraphics[width=0.375\linewidth]{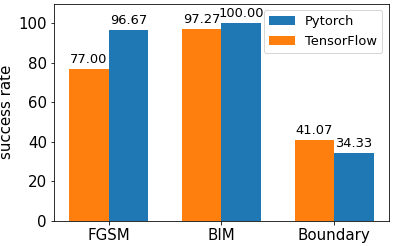}}
    \subfloat[ResNet50\label{gen_b}]{%
        \includegraphics[width=0.33\linewidth]{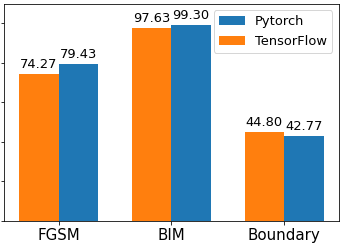}}
    \subfloat[InceptionV3\label{gen_c}]{%
        \includegraphics[width=0.33\linewidth]{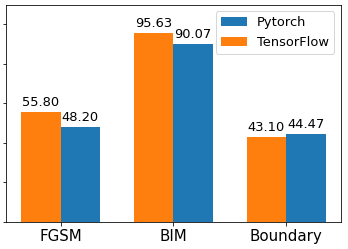}}
    \caption{Success rate of adversarial attacks on PC}
    \label{gen}
\end{figure}

\textbf{RQ2 results (TensorFlow and PyTorch robustness on PC vs. Mobile):}
To answer this RQ, we start by analyzing robustness on PC DL frameworks as our baseline. \figurename{~\ref{gen}} report the success rates of three adversarial attacks (FGSM, BIM, and Boundary) on three models (MobileNetV2, ReNet50, and InceptionV3) over two PC frameworks (TensorFlow and PyTorch). 

As it can be seen in \figurename{\ref{gen_a}} and \figurename{\ref{gen_b}} on MobileNetV2 and ResNet50 architectures TensorFlow is more robust against white-box attacks, while on Boundary attack which is black-box, PyTorch is more robust.
However, in InceptionV3 (\figurename{\ref{gen_c}}), it is the exact opposite, and TensorFlow is more robust against the black-box attack. Also, PyTorch is more robust against white-box attacks. These patterns were seen in our experiments on the mobile device (RQ1), as well.

\begin{table}[t]
    \setlength\tabcolsep{5pt}

    \caption{Adversarial generation time for the entire test sets (3,000 samples) reported in hours:minutes:seconds.}
    \label{gen_table}
    \centering
    \begin{tabular}{ccccc}
    \bottomrule
        &   &   \multicolumn{3}{c}{Generation time (s)}\\
        Model  &   Framework &   FGSM &   BIM  &   Boundary\\
        \toprule
        \multirow{2}{*}{MobileNetV2}    &   TensorFlow    &   00:02:20 &  00:04:13 & 01:01:46 \\
                                        &   PyTorch     &   00:03:35 & 00:04:16 & 01:26:04   \\
        \midrule
        \multirow{2}{*}{ResNet50}       &   TensorFlow    &   00:02:01 & 00:03:38 & 04:31:17\\
                                        &   PyTorch     & 00:02:30 &  00:05:37 & 04:56:24   \\
        \midrule
        \multirow{2}{*}{InceptionV3}    &   TensorFlow    &   00:03:06 &  00:03:45 &   05:54:14   \\
                                        &   PyTorch     &   00:01:36 & 00:05:18  & 07:25:36   \\
    \bottomrule
    \vspace{-0.06\linewidth}
    \end{tabular}
\end{table}

Table \ref{gen_table} reports  adversarial sample generation cost on PC. As it can be seen, the Boundary attack takes a significantly longer time to finish, with a much lower success rate in the end.
While white-box attacks finish in minutes, the Boundary attack takes a couple of hours to complete, even in a high-end system, such as Compute Canada cluster.
This perfectly illustrates why it is almost impossible to create samples using black-box techniques on mobile devices. 
Moreover, white-box attacks need gradient, which is unavailable on mobile frameworks at the moment. Thus they too cannot be run on mobile. Consequently, currently, there is no easy way to generate adversarial samples on mobile devices. Finally, we can see that TensorFlow is slightly faster than PyTorch, in most cases.

\begin{figure}[t]
    \vspace{-0.045\linewidth}

    \hspace{-0.045\linewidth}
    \subfloat[TensorFlow, MobileNetV2\label{platform_a}]{%
        \includegraphics[width=0.375\linewidth]{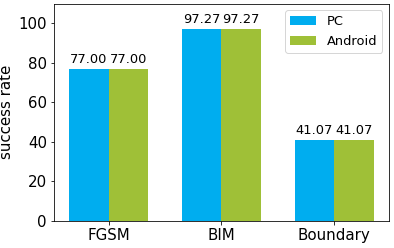}}
    \subfloat[TensorFlow, ResNet50\label{platform_b}]{%
        \includegraphics[width=0.33\linewidth]{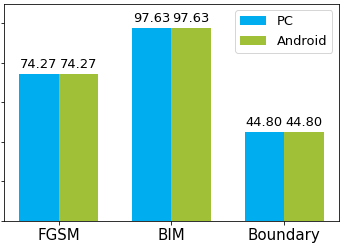}}
    \subfloat[TensorFlow, InceptionV3\label{platform_c}]{%
        \includegraphics[width=0.33\linewidth]{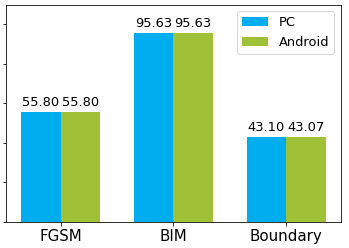}}
    \vfill
    \hspace{-0.045\linewidth}
    \subfloat[PyTorch, MobileNetV2\label{platform_d}]{%
        \includegraphics[width=0.375\linewidth]{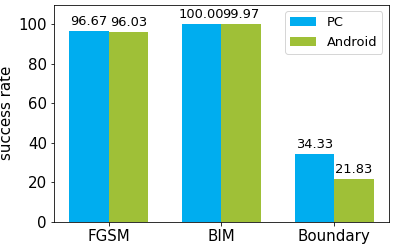}}
    \subfloat[PyTorch, ResNet50\label{platform_e}]{%
        \includegraphics[width=0.33\linewidth]{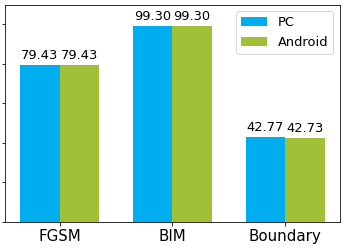}}
    \subfloat[PyTorch, InceptionV3\label{platform_f}]{%
        \includegraphics[width=0.33\linewidth]{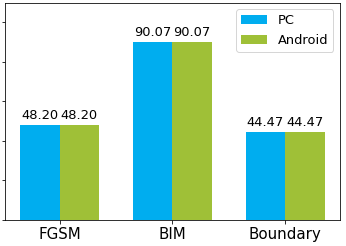}}
    \caption{Different attacks' success rates on different platforms.}
    \label{platform}
    \vspace{-0.04\linewidth}

\end{figure}

To better compare mobile and PC platforms, \figurename{\ref{platform}} reports the same raw data as Figures \ref{android_robustness} and \ref{gen}, but grouped by platforms.
As \figurename{\ref{platform_a}, \ref{platform_b}, and \ref{platform_c}} show, robustness is almost the same in all cases except on the Boundary attack on InceptionV3, where we see a slight increase in robustness on mobile. We also see similar patterns in \figurename{\ref{platform_d}, \ref{platform_e}, and \ref{platform_f}}, between mobile and PC for PyTorch where the robustness on mobile is either the same or very close to the PC version in all configurations. The exception is the Boundary attack on MobileNetV2, where we see a sudden drop in success rate in PyTorch Mobile.
This shows PyTorch's mobile version of MobileNetV2 is more robust against the Boundary attack.

\textbf{Answer to RQ2:}  
In most cases, switching platforms between PC and mobile does not change the robustness drastically.
This means that the implementation of models on both hardware and languages perform similarly, and they are almost equivalent.
However, in some cases, like on PyTorch when using Boundary attack on MobileNetV2, we might get much higher robustness on mobile platforms.

\begin{figure}[t]
    \centering
    \hspace{-0.045\linewidth}
    \subfloat[TensorFlow, MobileNetV2\label{quant_a}]{%
        \includegraphics[width=0.375\linewidth]{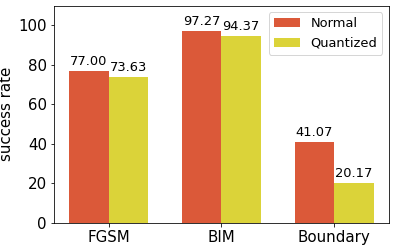}}
    \subfloat[TensorFlow, ResNet50\label{quant_b}]{%
        \includegraphics[width=0.33\linewidth]{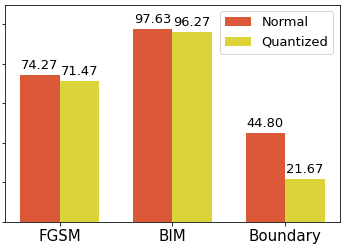}}
    \subfloat[TensorFlow, InceptionV3\label{quant_c}]{%
        \includegraphics[width=0.33\linewidth]{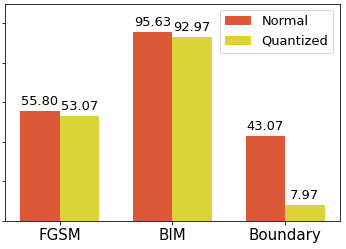}}
    \vfill
    \hspace{-0.045\linewidth}
    \subfloat[PyTorch, MobileNetV2\label{quant_d}]{%
        \includegraphics[width=0.375\linewidth]{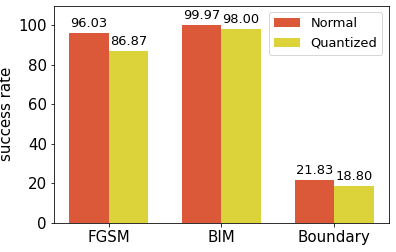}}
    \subfloat[PyTorch, ResNet50\label{quant_e}]{%
        \includegraphics[width=0.33\linewidth]{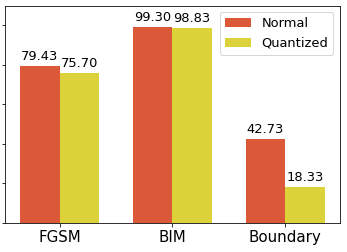}}
    \subfloat[PyTorch, InceptionV3\label{quant_f}]{%
        \includegraphics[width=0.33\linewidth]{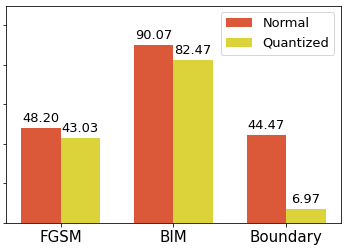}}
    \caption{Effect of quantization on success rate}
    \label{quant}

\end{figure}

\textbf{RQ3 results (the quantization effect):}
To answer this RQ, we report the results in \figurename{\ref{quant}}.
The results show that the attacks lose their initial effectiveness in all cases and their success rate decreases. 
This shows that quantization can increase the robustness of the model against attacks. 
In some cases, robustness increases slightly, whereas in a case like a Boundary attack (which is trying to create samples closest to the boundary), the slightest effect like quantization can massively improve the robustness. In other words, these attacks are very dependent on the model's specifications.
If model parameters change (as quantization does), the attack will not be as effective as it was.

The median decrease of success rates per attack is 3.55\% (FGSM), 2.43\% (BIM), and 23.77\% (Boundary), with a minimum of 0.47\% (for BIM om PyTorch-ResNet50) and a maximum of 37.5\% (For Boundary on PyTorch-InceptionV3). This difference between the unquantized and quantized models' robustness is statistically significant with a p-value less than 0.001 when running a non-parametric statistical significant test (Paired Wilcoxon Signed-Rank test), with the effect size measure (Paired Vargha and Delaney A Measure) is 0.608.

In terms of model performance, as Table \ref{quantPerformance} shows, quantization closes the model inference time gap between the two frameworks, and in some models like MobileNetV2, PyTorch Mobile even runs faster than TensorFlow Lite.
As expected, the speedup after quantization is significant, and it goes up to 2.87 times in some cases (e.g., MobileNetV2 in PyTorch).

\begin{table}
    \setlength\tabcolsep{5pt}

    \caption{Effect of quantization on mobile inference time}
    \label{quantPerformance}
    \centering
    \begin{tabular}{ccccc}
        \bottomrule
        &   &   &   \multicolumn{2}{c}{Inference time (s)}   \\
        Model & Attack & Framework & Regular & Quantized\\
        \toprule
        \multirow{6}{*}{MobileNetV2}	&	\multirow{2}{*}{FGSM}	&	TensorFlow Lite	&	157 & 133\\
        	&		&	PyTorch Mobile	&	276 & 96\\
        \cline{2-5}	&	\multirow{2}{*}{BIM}	&	TensorFlow Lite	&	174 & 128\\
        	&		&	PyTorch Mobile	&	301 & 105\\
        \cline{2-5}	&	\multirow{2}{*}{Boundary}	&	TensorFlow Lite	&	187 & 132\\
        	&		&	PyTorch Mobile	&	302 & 112\\
        \midrule
        \multirow{6}{*}{ResNet50}	&	\multirow{2}{*}{FGSM}	&	TensorFlow Lite	&	1003 & 534\\
        	&		&	PyTorch Mobile	&	1583 & 616\\
        \cline{2-5}	&	\multirow{2}{*}{BIM}	&	TensorFlow Lite	&	1036 & 531\\
        	&		&	PyTorch Mobile	&	1504 & 622\\
        \cline{2-5}	&	\multirow{2}{*}{Boundary}	&	TensorFlow Lite	&	1093 & 559\\
        	&		&	PyTorch Mobile	&	1460 & 554\\
        \midrule
        \multirow{6}{*}{InceptionV3}	&	\multirow{2}{*}{FGSM}	&	TensorFlow Lite	&	1617 & 798\\
        	&		&	PyTorch Mobile	&	1774 & 942\\
        \cline{2-5}	&	\multirow{2}{*}{BIM}	&	TensorFlow Lite	&	1626 & 804\\
        	&		&	PyTorch Mobile	&	1841 & 982\\
        \cline{2-5}	&	\multirow{2}{*}{Boundary}	&	TensorFlow Lite	&	1516 & 808\\
        	&		&	PyTorch Mobile	&	1820 & 993\\
        	\bottomrule
    \end{tabular}

\end{table}

\textbf{Answer to RQ3:} In addition to the speed and size reduction that quantization provides, it can be a very low-cost and straightforward defense mechanism against adversarial attacks. Quantization increases the robustness with a median up to 37.5\% for some attacks (Boundary) over a distribution of three models and two DL frameworks per attack.

\subsection{Threats to validity}
In terms of the construct validity, the key to our success rate measure is knowing the ground truth. However, we are only relying on the original classification models to come up with the ground truth. That is, there might be cases where the model is misclassifying, but the adversarial sample is correctly classified. As discussed, we take the intersection of the three models' correctly classified samples to reduce the probability of having a misclassified ground truth. 

In terms of the internal validity, our study design is pretty simple, with evaluating the robustness of models using attacks success rate. We are using predefined classification models and existing libraries to create attacks. Therefore, we are not introducing confounding factors in our implementation or design.   

In terms of conclusion validity and to address the randomness of the results, we ran a non-parametric paired statistical significant test (Paired Wilcoxon Signed-Rank test) and reported the effect size measure (Paired Vargha and Delaney A Measure) when comparing the success rates in RQ3. In RQ1 and RQ2, the results were mainly the same when comparing frameworks and platforms. Therefore, our conclusions were the differences are practically insignificant anyways, so there was no need to run any statistical significance tests.  

Finally, regarding the external validity and generalizability of the study, one potential validity threat is having limited datasets and models, which results in biased conclusions.
We mitigate this threat by choosing one of the most extensive datasets in the image domain.
Furthermore, we used three state-of-the-art models with different complexities to help generalize our results.
Finally, we used both white-box and black-box attacks, and all the attacks were some of the best in their categories.
However, still, all of these datasets and models are from the domain of the image. 
Thus the results might not be generalizable to other domains such as natural language processing.
In addition, we only used Android as our mobile platform. 
Therefore, the results might not be representative of iOS.

\section{Related Work}
\label{relSec}
Luo et al. \cite{DBLP:journals/corr/abs-2005-05085} made a comparison for classifiers between different mobile frameworks like TensorFlow Lite PyTorch Mobile and Caffe 2 \cite{Caffe2AN39:online}, which is now part of PyTorch.
They used many models such as ResNet50, InceptionV3, DenseNet121 \cite{huang2018densely}, and compared all the models on all the mentioned frameworks. 
They also compared the neural inference power of different mobile devices. 
Some of the results were as follows: none of the platforms had a noticeable advantage in all cases, TensorFlow Lite had a much faster model loading time compared to the others, the same AI model on the different platform had different accuracy, and Android Neural Networks API (NNAPI) did not constantly improve the inference time.
This study was mainly focused on accuracy and performance but did not have any robustness assessments, which our study covers.

Ignatov et al. \cite{DBLP:conf/eccv/IgnatovTCWWHG18} made a benchmark consisting of multiple tests such as image recognition, face recognition, image deblurring, image semantic segmentation, image enhancement, and memory limitations.
Then they compared the performance of DNN models on different mobile phones.
This paper's main idea was to measure the power of the CPU chipset; thus, there was no other comparison in this work.
They only used TensorFlow Lite as the DL framework, and there was no comprehensive study on the impact of different models.


Guo et al. \cite{DBLP:conf/kbse/GuoCXMHLLZL19} presented a study on PyTorch Mobile, TensorFlow Lite and
TensorFlow.js\cite{TensorFl14:online}, CNTK\cite{DBLP:conf/kdd/SeideA16}, MXnet\cite{DBLP:journals/corr/ChenLLLWWXXZZ15}.
The paper made a comparison on PC and found that PyTorch and MXnet were more vulnerable in adversarial attacks.
It compared browsers using TensorFlow.js with PC for MNIST and CIFAR-10 datasets using different models and found that TensorFlow.js suffered from high memory usage and had meaningful lower accuracy ResNet model.
Android devices were faster in small models in inference time, whereas IOS devices were better at large models.
It used TensorFlow Lite to compare Android and iOS devices with PC and found similar accuracy to PC.
Finally, they found that quantization did not affect accuracy much, and it made inference faster on Android devices. Although there was some robustness analysis in this work, the models were very simple and unrealistic, and evaluation was only on PC.

Huang et al. \cite{seip} made some interesting experiments on the robustness of models on Android devices. 
They used TensorFlow Lite as the framework for their study. Their approach had some key points. They extracted TensorFlow Lite models from the Google Play store. Then based on some criteria, They found similar pre-trained PC models available online and implemented the attacks on similar models.
Their results showed that their approach was more effective in fooling the models than blind attacks (attacks without knowing the model).
However, this study was only focused on attacking a specific model on mobile, and It did not look at different frameworks and the effect of platforms.

Given the related work, we see a gap in the literature for assessing the robustness of mobile DL frameworks, which our study covers. 

\section{Acknowledgement}
This work was enabled in part by support from WestGrid (www.westgrid.ca) and Compute Canada (www.computecanada.ca) and the Natural Sciences and Engineering Research Council of Canada [RGPIN/04552-2020].

\section{Conclusion and Future works}
\label{conSec}
In this paper, we conduct a comprehensive study on deep learning mobile frameworks' robustness with different configurations.
We compare the two major mobile frameworks (TensorFlow Lite and PyTorch Mobile), using 18 configurations (36 configurations considering quantization): two frameworks, three models, and three adversarial attack techniques.
Our results show that frameworks are not necessarily superior in terms of robustness on the mobile platform, and the more robust framework varies by model architecture and attack type. Furthermore, changing the platform to mobile usually does not affect robustness but in some cases results in a slight increase in robustness which is not significant.
However, we also show that quantization is a very effective approach in reducing the cost of model inference and making it more robust toward attacks in DL frameworks, consistently improving the robustness of Mobile DL frameworks (even up to 37.5\% improvement when compared to regular models). 
In the future, we plan to extend this study to other application domains (such as textual data) and study other frameworks and platforms such as TensorFlow.js and iOS devices. 

\newpage
\bibliographystyle{splncs04}    
\bibliography{refs}

\begin{thebibliography}{10}
\providecommand{\url}[1]{\texttt{#1}}
\providecommand{\urlprefix}{URL }
\providecommand{\doi}[1]{https://doi.org/#1}

\bibitem{Caffe2AN39:online}
Caffe2 | a new lightweight, modular, and scalable deep learning framework.
  https://caffe2.ai/

\bibitem{HomePyTo69:online}
Home | pytorch. https://pytorch.org/mobile/home/

\bibitem{ImageNet5:online}
Imagenet. http://www.image-net.org/

\bibitem{Inceptio34:online}
Inception\_v3 | pytorch.
  https://pytorch.org/hub/pytorch\_vision\_inception\_v3/

\bibitem{Introduc74:online}
Introduction to quantization on pytorch | pytorch.
  https://pytorch.org/blog/introduction-to-quantization-on-pytorch/

\bibitem{KerasApp64:online}
Keras applications. https://keras.io/api/applications/

\bibitem{kerasapp55:online}
keras-applications/imagenet\_utils.py at 1.0.8.
  https://github.com/keras-team/keras-applications/blob/1.0.8/keras\_applications/imagenet\_utils.py

\bibitem{Modelopt44:online}
Model optimization | tensorflow lite.\\
  https://www.tensorflow.org/lite/performance/model\_optimization

\bibitem{TensorFl80:online}
Tensorflow lite | ml for mobile and edge devices.
  https://www.tensorflow.org/lite

\bibitem{TensorFl14:online}
Tensorflow.js | machine learning for javascript developers.
  https://www.tensorflow.org/js

\bibitem{torchvis33:online}
torchvision — pytorch 1.7.0 documentation.\\
  https://pytorch.org/docs/stable/torchvision/index.html

\bibitem{torchvis36:online}
torchvision.models — torchvision master documentation.
  https://pytorch.org/vision/stable/models.html

\bibitem{WordNetA96:online}
Wordnet | a lexical database for english. https://wordnet.princeton.edu/

\bibitem{10.5555/3026877.3026899}
Abadi, M., Barham, P., Chen, J., Chen, Z., Davis, A., Dean, J., Devin, M.,
  Ghemawat, S., Irving, G., Isard, M., Kudlur, M., Levenberg, J., Monga, R.,
  Moore, S., Murray, D.G., Steiner, B., Tucker, P., Vasudevan, V., Warden, P.,
  Wicke, M., Yu, Y., Zheng, X.: Tensorflow: A system for large-scale machine
  learning. In: Proceedings of the 12th USENIX Conference on Operating Systems
  Design and Implementation. p. 265–283. OSDI'16, USENIX Association, USA
  (2016)

\bibitem{DBLP:conf/iclr/BrendelRB18}
Brendel, W., Rauber, J., Bethge, M.: Decision-based adversarial attacks:
  Reliable attacks against black-box machine learning models (2018)

\bibitem{7958570}
{Carlini}, N., {Wagner}, D.: Towards evaluating the robustness of neural
  networks. In: 2017 IEEE Symposium on Security and Privacy (SP). pp. 39--57
  (2017). \doi{10.1109/SP.2017.49}

\bibitem{DBLP:journals/corr/ChenLLLWWXXZZ15}
Chen, T., Li, M., Li, Y., Lin, M., Wang, N., Wang, M., Xiao, T., Xu, B., Zhang,
  C., Zhang, Z.: Mxnet: A flexible and efficient machine learning library for
  heterogeneous distributed systems (2015)

\bibitem{10.5555/3157096.3157279}
Fawzi, A., Moosavi-Dezfooli, S.M., Frossard, P.: Robustness of classifiers:
  From adversarial to random noise. In: Proceedings of the 30th International
  Conference on Neural Information Processing Systems. p. 1632–1640. NIPS'16,
  Curran Associates Inc., Red Hook, NY, USA (2016)

\bibitem{DBLP:journals/corr/GoodfellowSS14}
Goodfellow, I.J., Shlens, J., Szegedy, C.: Explaining and harnessing
  adversarial examples. In: Bengio, Y., LeCun, Y. (eds.) 3rd International
  Conference on Learning Representations, {ICLR} 2015, San Diego, CA, USA, May
  7-9, 2015, Conference Track Proceedings (2015),
  http://arxiv.org/abs/1412.6572

\bibitem{DBLP:conf/kbse/GuoCXMHLLZL19}
Guo, Q., Chen, S., Xie, X., Ma, L., Hu, Q., Liu, H., Liu, Y., Zhao, J., Li, X.:
  An empirical study towards characterizing deep learning development and
  deployment across different frameworks and platforms. In: 34th {IEEE/ACM}
  International Conference on Automated Software Engineering, {ASE} 2019, San
  Diego, CA, USA, November 11-15, 2019. pp. 810--822. {IEEE} (2019).
  \doi{10.1109/ASE.2019.00080}, https://doi.org/10.1109/ASE.2019.00080

\bibitem{DBLP:conf/cvpr/HeZRS16}
He, K., Zhang, X., Ren, S., Sun, J.: Deep residual learning for image
  recognition. In: 2016 {IEEE} Conference on Computer Vision and Pattern
  Recognition, {CVPR} 2016, Las Vegas, NV, USA, June 27-30, 2016. pp. 770--778.
  {IEEE} Computer Society (2016). \doi{10.1109/CVPR.2016.90},
  https://doi.org/10.1109/CVPR.2016.90

\bibitem{huang2018densely}
Huang, G., Liu, Z., van~der Maaten, L., Weinberger, K.Q.: Densely connected
  convolutional networks (2018)

\bibitem{seip}
Huang, Y., Hu, H., Chen, C.: Robustness of on-device models: Adversarial attack
  to deep learning models on android apps (2021)

\bibitem{DBLP:conf/eccv/IgnatovTCWWHG18}
Ignatov, A., Timofte, R., Chou, W., Wang, K., Wu, M., Hartley, T., Gool, L.V.:
  Ai benchmark: Running deep neural networks on android smartphones. In:
  Leal-Taix{\'e}, L., Roth, S. (eds.) Computer Vision -- ECCV 2018 Workshops.
  pp. 288--314. Springer International Publishing, Cham (2019)

\bibitem{jacob2017quantization}
Jacob, B., Kligys, S., Chen, B., Zhu, M., Tang, M., Howard, A., Adam, H.,
  Kalenichenko, D.: Quantization and training of neural networks for efficient
  integer-arithmetic-only inference (2017)

\bibitem{kurakin2017adversarial}
Kurakin, A., Goodfellow, I., Bengio, S.: Adversarial examples in the physical
  world (2017)

\bibitem{DBLP:journals/corr/abs-2005-05085}
Luo, C., He, X., Zhan, J., Wang, L., Gao, W., Dai, J.: Comparison and
  benchmarking of ai models and frameworks on mobile devices (2020)

\bibitem{7780651}
Moosavi-Dezfooli, S.M., Fawzi, A., Frossard, P.: Deepfool: A simple and
  accurate method to fool deep neural networks. In: 2016 IEEE Conference on
  Computer Vision and Pattern Recognition (CVPR). pp. 2574--2582 (2016).
  \doi{10.1109/CVPR.2016.282}

\bibitem{paszke2017automatic}
Paszke, A., Gross, S., Chintala, S., Chanan, G., Yang, E., DeVito, Z., Lin, Z.,
  Desmaison, A., Antiga, L., Lerer, A.: Automatic differentiation in pytorch
  (2017)

\bibitem{rauber2017foolboxnative}
Rauber, J., Zimmermann, R., Bethge, M., Brendel, W.: Foolbox native: Fast
  adversarial attacks to benchmark the robustness of machine learning models in
  pytorch, tensorflow, and jax. Journal of Open Source Software  5(53), ~2607
  (2020). \doi{10.21105/joss.02607}

\bibitem{DBLP:conf/cvpr/SandlerHZZC18}
Sandler, M., Howard, A.G., Zhu, M., Zhmoginov, A., Chen, L.: Mobilenetv2:
  Inverted residuals and linear bottlenecks. In: 2018 {IEEE} Conference on
  Computer Vision and Pattern Recognition, {CVPR} 2018, Salt Lake City, UT,
  USA, June 18-22, 2018. pp. 4510--4520. {IEEE} Computer Society (2018).
  \doi{10.1109/CVPR.2018.00474}

\bibitem{DBLP:conf/kdd/SeideA16}
Seide, F., Agarwal, A.: {CNTK:} microsoft's open-source deep-learning toolkit.
  In: Krishnapuram, B., Shah, M., Smola, A.J., Aggarwal, C.C., Shen, D.,
  Rastogi, R. (eds.) Proceedings of the 22nd {ACM} {SIGKDD} International
  Conference on Knowledge Discovery and Data Mining, San Francisco, CA, USA,
  August 13-17, 2016. p.~2135. {ACM} (2016). \doi{10.1145/2939672.2945397}

\bibitem{8601309}
{Su}, J., {Vargas}, D.V., {Sakurai}, K.: One pixel attack for fooling deep
  neural networks. IEEE Transactions on Evolutionary Computation  23(5),
  828--841 (2019). \doi{10.1109/TEVC.2019.2890858}

\bibitem{DBLP:conf/cvpr/SzegedyVISW16}
Szegedy, C., Vanhoucke, V., Ioffe, S., Shlens, J., Wojna, Z.: Rethinking the
  inception architecture for computer vision. In: 2016 {IEEE} Conference on
  Computer Vision and Pattern Recognition, {CVPR} 2016, Las Vegas, NV, USA,
  June 27-30, 2016. pp. 2818--2826. {IEEE} Computer Society (2016).
  \doi{10.1109/CVPR.2016.308}

\bibitem{DBLP:journals/corr/abs-2004-09602}
Wu, H., Judd, P., Zhang, X., Isaev, M., Micikevicius, P.: Integer quantization
  for deep learning inference: Principles and empirical evaluation (2020)

\end{thebibliography}
\end{document}